# A Survey on State-of-the-art Techniques for Knowledge Graphs Construction and Challenges ahead

Ali Hur
School of Science, Computing, and Security, Edith Cowan University
Perth, Australia
mahur@our.ecu.edu.au

Naeem Janjua[1,2]
[1]School of Science, Computing, and Security, Edith Cowan University, Perth, Australia; [2]Faculty of Computing and AI, Air University, Islamabad, Pakistan.
n.janjua@ecu.edu.au; naeem.janjua@au.edu.pk

Mohiuddin Ahmed
School of Science, Computing, and Security, Edith Cowan University
Perth, Australia
mohiuddin.ahmed@ecu.edu.au

*Abstract* -- Global datasphere is increasing fast, and it is expected to reach 175 Zettabytes by 2025[1]. However, most of the content is unstructured and is not understandable by machines. Structuring this data into a knowledge graph enables multitudes of intelligent applications such as deep question answering, recommendation systems, semantic search, etc. The knowledge graph is an emerging technology that allows logical reasoning and uncovers new insights using content along with the context. Thereby, it provides necessary syntax and reasoning semantics that enable machines to solve complex healthcare, security, financial institutions, economics, and business problems. As an outcome, enterprises are putting their effort into constructing and maintaining knowledge graphs to support various downstream applications. Manual approaches are too expensive. Automated schemes can reduce the cost of building knowledge graphs up to 15-250 times. This paper critiques state-of-the-art automated techniques to produce knowledge graphs of near-human quality autonomously. Additionally, it highlights different research issues that need to be addressed to deliver high-quality knowledge graphs.

## I. Introduction

A knowledge graph is a data structure that represents knowledge in the form of nodes and edges. Nodes represent descriptions about real-world entities, and edges represent the associations between these entities. Knowledge graph provides a simple and convenient way of representing machine-readable knowledge. It can express rich semantics and support advanced reasoning, which empowers applications like question-answering, semantic search, recommendations, language understanding, and advanced analytics. Knowledge graph plays a vital role in the development of such kinds of applications [2]. It provides annotations of the data items and their conceptual relationships that enable complex reasoning over the facts to derive new information. For instance, a semantic search application needs to determine the intent and semantic meaning of the keywords to answer search queries with high accuracy. Knowledge graph serves as a richly structured data source that semantic search applications can use to provide highly contextualized and accurate results. It establishes insights into the business realm and enables advanced reasoning to discover unseen conceptual connections, even between heterogeneous data items. For example, inductive learning techniques or graph-based analytics can be performed to learn useful functional patterns or rules that can predict possible outcomes. Thus, it plays a vital role in implementing robust recommendations systems. Adaptive and personalized educational systems are other examples of applications that discover semantic relations between user characteristics and other domain objects to provide personalized suggestions.

Knowledge graph enables structuring of information to represent knowledge. It also provides a convenient way to add new links and even extend entities with existing knowledge sources like Conceptnet or Wikidata. It delivers the best platform for applications in conversational AI (like chatbots and intelligent assistants), which require proper interpretation of context to understand the meanings of terms and intents in user conversations. In semantic web standards, knowledge representation languages like RDF/OWL/SPARQL, are based on the same underlying data structure and have close relationships with knowledge graphs [1]. The term knowledge graph was coined by Google in 2012 while releasing its knowledge graph, which is also known as knowledge vault [2]. Some examples of existing large knowledge graphs include DBpedia, Wikidata, Freebase, NELL, and YAGO.

Big enterprises construct and use knowledge graphs to support their knowledge requirements and enable intelligent downstream applications [3]. Knowledge graphs are built either manually or in some automated manner. Manual approaches like CYC [4] rely on the services of highly qualified domain experts and knowledge engineers. The other approach of manually constructing knowledge graphs is to rely on communities or interest groups by organizing crowdsourcing efforts. Conceptnet [5], Wikidata [6], Freebase [7], and [8] are some real-world examples of crowdsourced knowledge graphs. Manual construction is labor-intensive, carries human biases, is tough, and is prone to human errors. Therefore, over the last two decades, the research community has been focusing on devising and implementing automated methods to build knowledge graphs [9], [10], [11]. The last decade has seen significant achievements in Knowledge harvesting and information extraction. Most of the content on the internet is unstructured (natural language text) and semi-structured (tables, trees, spreadsheets), which provides a rich source of available knowledge. Due to the abundance of publicly accessible information, substantial efforts such as

---

[1] For detailed statistics see: https://www.seagate.com/files/www-content/our-story/trends/files/idc-seagate-dataage-whitepaper.pdf

[2] https://allegrograph.com/gartner-hype-cycle-for-ai-knowledge-graphs/

[15], [16], [17], [18] have been made to collect or harvest knowledge from different sources over the internet. Techniques like text mining and natural language processing have been proposed to extract information from these sources in triples. The outcome is an extraction graph [12] or a data graph [11].

These extraction techniques can collect candidate facts in the form of entities, their attributes, and relations between these entities. Nevertheless, due to the noisy nature of these sources, the resulting graph contains a lot of redundant, invalid, and inconsistent facts [13]. Moreover, it lacks necessary background knowledge, semantic descriptions of entities, and relationships, and therefore, is not considered sufficient for enterprises' knowledge requirements [14]. According to [11], 99% of existing knowledge graphs are data graphs. Moreover, in another survey [15], the authors have figured out that only 19% of the triples in the NELL-995 conform to the NELL schema. Manually removing these errors is too costly and time-consuming. For example, NELL uses periodic human supervision to detect and alleviate incorrect triples, which becomes a costly and time-consuming effort. Therefore, some automated correction mechanisms should be in place [16]. In the case of enterprise-level or industrial scale knowledge graphs, identifying and fixing invalid or inconsistent information is considered a mandatory requirement [17]. Automated correctness schemes have been proposed to identify and fix these errors that we will review in the forthcoming sections of this paper.

Coverage and correctness are two significant challenges even for large knowledge graphs [25][24]. Similarly, a lot of missing information needs to be revealed to make the knowledge graph complete. According to a study mentioned in [18], almost half of DBPedia [3] entities contain less than five relationships. Another study reveals that the birthplace of over 70% of people in Freebase is unknown, and 90% have no mention of ethnicity [19]. In response to such issues, knowledge graph refinement techniques have been proposed in the literature. This paper focuses on surveying different categories of automated techniques that can improve the quality of the knowledge graph in terms of its overall coverage and correctness. Furthermore, it will highlight each category's pros and cons and discuss the associated research issues and challenges.

The rest of this paper is organized as follows. Section 2 gives a general overview of the knowledge graph construction pipeline and the background of knowledge graph refinement activities: section 3 and 4 review different categories of knowledge graph completion and correction techniques. Section 5 summarizes the study of existing techniques and sheds light on emerging research issues and challenges. Lastly, section 6 concludes the discussion.

## II. OVERVIEW

In this section, we first introduce the background of the knowledge graph construction pipeline and then give a general overview of knowledge graph refinement activities, their goals, and objectives.

### A. Knowledge graph construction pipeline

The knowledge graph construction pipeline is a set of components executed in sequence to construct a knowledge graph. It can also be thought of as construction stages, giving information to the subsequent phases. A typical knowledge graph construction pipeline can be conceptually divided into five phases. These include specification, conceptualization, formalization, integration, and augmentation. The specification phase states the overall motivation behind building the knowledge graph, its purpose, and scope.

Once needs have been identified and analyzed, the next step is to properly document them in ORSD (ontology requirement specification document). ORSD is usually written in natural language using a set of intermediate representations to specify business constraints and requirements, use case descriptions, the overall scope of the project, glossary of domain terms, a grouping of related terms, and a set of CQs (competency questions). Moreover, it asserts domain specifications and a set of functional and non-functional requirements.

After the specification phase, the next phase is the conceptualization of the domain. The core objective of this phase is to give structure to already collected domain knowledge in the form of "intermediate representations." This phase describes the problem and its solution in terms of domain vocabularies, including identifying data schemata, attributes, classes, and links connecting different data schemas. The outcome will be a high-level conceptual representation that contains main concepts and associations among them. They serve as foundations with necessary and minimal information to enable subsequent growth of the knowledge graph. The conceptual model of the domain is then transformed into a formal model. This formalization phase is concerned with the transformation of the resulting conceptual model into a standard machine-interpretable model. Different formal languages like OWL, Description Logic, Framenets, RDFS, etc., can formalize the conceptual model. The outcome will be a formal model of the domain objects, their relationships, and rules and constraints. Qualified knowledge engineers and domain experts produce this formal model.

The formal model is then extended by integrating existing knowledge sources. These sources are usually identified in the specification stage of the KG construction pipeline. In most cases, the relevant part of the existing knowledge source will be integrated using entity and schema alignment techniques.

All these phases described above establish the domain specifications, the schema, standard vocabularies, and integration of existing datasets to create a fact base for knowledge graphs. The next phase, i.e., augmentation, is related to knowledge acquisition (such as schema learning and population) and refinement (KG completeness and correctness). In the augmentation phase, we have three things to cover; the first focuses on the enrichment and expansion of the knowledge graph. The extractors are used to extract information from text, HTML documents, DOM elements, or human-annotated elements. The second one assesses the quality of the knowledge graph to determine whether it can be used reliably [20]. The objective is to devise refinement strategies to improve the quality. Finally, the third one deals with the refinement of the knowledge graph, which discovers hidden links and removes anomalies. We will see in the next section further detail about different types of refinement techniques and approaches to implement them.

### B. knowledge graph refinement

The main goal is to improve the quality of the knowledge graph so that it can be reliably used and fulfill the business requirements set by an organization. Different refinement approaches encompassing techniques like rule/pattern mining, probabilistic graphical models, distributed representation, and recently deep neural network models have been proposed. Each one has its strengths and weaknesses. For example, neural network-based approaches perform well in extracting high-quality triples but only in minimal numbers. In contrast, the OpenIE based methods tend to find many triples, but few of them are of high quality [21]. Therefore, refinement strategies need to be devised to remove the anomalies like redundancy, incorrect and contradictory assertions, and discover hidden relationships within entities in an incomplete knowledge graph. These refinement techniques can be divided into KG completion and KG correction techniques.

*C. KG Completeness*

The objective of the knowledge graph completeness activity is to increase the overall coverage of the knowledge graph. Coverage implies that all the required information has been integrated into the knowledge graph related to the domain, and it supports all targeted use cases [17]. The idea is to complement hidden facts within the existing content of the knowledge graph [22]. Various techniques of link predictions have been proposed to find different types of links. These links can be classified into three broad categories, i.e., general links, identity links, and type links [20].

General links refer to connections between nodes of arbitrary labels (e.g., relations between individuals). These connections are usually inferred using techniques like inductive learning, knowledge graph embeddings, and logical rule/axiom mining. Type links represent the connections between entities and their types, e.g., typeOf relation in OWL/RDFS. It is related to the classification problem, e.g., a classifier can be trained on the features of nodes by inspecting their inward and outward edges to infer node type. Identity links indicate connections between similar entities. For example., sameAs relation in OWL/RDFS. The process of identifying identity links between entities is also known as deduplication, record linkage, or entity resolution, where some similarity matching technique is employed. These matching techniques can be divided into two groups 1. Value matchers – that involve similarity measures on the value of the properties of the nodes, e.g., string, number, date, etc. 2. Context matchers – measure similarity based on surrounding information.

*D. KG Correction*

Similarly, in contrast to completion, whose main objective is to discover new/hidden facts, correction refers to identifying and removing erroneous facts. Two main approaches to correcting the knowledge graph are fact validation and inconsistency repairs [20]. Fact validation refers to the assessment of the facts to assign plausibility scores between 1 and 0. It is also known as fact-checking [23]. The objective is to analyze the statements in the knowledge graph to check whether they are semantically correct and conforms to the real world [23]. Another challenge is to validate the facts considering their temporal values [24]. Inconsistency repairs mean that the facts in the knowledge graph should adhere to the axioms/ restrictions defined in the ontologies [25]. For example, not adhering to the Disjointness axiom may cause inconsistency in a knowledge graph. Therefore, it is a two-step process, where the first one focuses on detecting the inconsistencies and the second one deals with repairing these inconsistencies.

III. KG COMPLETENESS APPROACHES

*A. Rule Mining and Pattern Mining*

After extraction, the knowledge graph is mainly populated with extensional knowledge (i.e., statements or facts) but without logical invariants. These rational invariants in the form of rules and constraints represent the intensional knowledge [26]. Rules and constraints play an essential role in refining knowledge graphs and enabling application use-cases like query answering and reasoning. As far as their role in KG refinement is concerned, rules help in deriving new assertions/facts via deductive reasoning and play a vital role in knowledge graph construction [27][21] and completion [2]. Rule mining and pattern mining approaches tend to capture hidden patterns in the underlying graph that conveniently and significantly improve the accuracy and completeness of knowledge graphs [28][16].

Rules need to be either manually created or automatically by analyzing the facts in the knowledge graph. Manual approaches are error-prone, cumbersome, and time-consuming. Therefore much focus of knowledge refinement research is on rule mining/pattern mining techniques. These techniques are employed to automatically detect rules from the existing facts in the knowledge graphs. Logical Rules can be expressed in chosen fragments of first-order predicate logic where full expressiveness is given up, minimizing computational complexity. For example, the simplest form of deduction rules can be represented in RDFS (Resource Description Framework Schema) language, which allows domain and range, subclass, and sub-property types of rules. For more advanced rules like disjointness, more expressive language like OWL (Web Ontology Language) is used. OWL is based on description logic which is a chosen set of fragments of first-order predicate logic.

Rule mining or frequent pattern mining exploits the techniques of inductive logic programming[29] and association rule mining [30]. These techniques identify interesting logical patterns within the facts of the knowledge graph, which can later assist in inferring new facts. Simple Rules can be learned through techniques like association rule mining [30] or Apriori rule mining, but these rules can only express single-variable patterns. Therefore, these rules cannot reach full logical expressivity and are limited to simple KG refinement tasks. Inductive logic programming techniques can learn more expressive rules (also known as horn rules). The patterns can be extracted from structured, semi-structured, and unstructured content. A deductive rule-based inference mechanism can be applied to discover missing links and associations between different entities. Consequently, it helps in filling the gaps in the knowledge graph.

Besides, rules and constraints play a crucial role in knowledge graph completeness and correctness; they also have other powerful features like interpretability and explainability, making them suitable to sensitive domains like medicine and drug discovery. Each rule has its support and confidence score associated that is assigned during the rule learning process. Rules that have high value for support and confidence are considered. The others are discarded.

Rule mining approaches usually suffer from limitations like directionality and localness [22]. Directionality means that the detected pattern assumes causality (i.e., based on a very restrictive assumption that co-occurring patterns are due to causality). Whereas localness implies that the pattern detected only affects the neighborhood nodes and ignores its global impact (i.e., the effect of executing the rule only appears on the neighborhood nodes). In addition to these issues, rule mining approaches are not considered scalable, particularly in large knowledge graphs.

Some work has been done on the scalability issue of rule mining/ pattern mining approaches. For example, ScaleKB [31] proposed a scalable system to mine first-order inference rules using an ontological pathfinding algorithm. They applied their technique on Freebase and surpassed many state-of-the-arts methods. The limitation of this approach is that it only focuses on static content and does not consider dynamic knowledge bases where each update in KG requires rerunning the learning algorithm, which is practically unfeasible. Therefore, the existing approaches need to be extended to perform online learning with expanding knowledge bases. In Knowledge Vault [2], the authors have proposed a scheme that uses the PRA (path ranking algorithm) to mine rules like association rule mining. This scheme uses an external knowledge source as a Freebase [7] to mine patterns that help predict new links in the knowledge graph. It can use the existing knowledge base (what they call graph prior) to analyze the current edges and then using schemes like PRA and neural network model (such as MLP) based approaches to infer new facts. See [2] for more details. In [28], the authors have proposed a rule mining system that reveals generic and conditional rules from inconsistent and incomplete knowledge graphs. Their system can generate both positive and negative rules to complete and correct the knowledge graph.

*B. Statistical and Probabilistic approaches*

Statistical and Probabilistic approaches [32] nicely deal with the notion of uncertainty and enable powerful reasoning ability just like humans do. They derive conclusions about a particular situation using inductive learning techniques where facts participating in the context are not deterministic but uncertain and carry a certain probability or likelihood. The facts have their likelihood value which is calculated based on some probabilistic features. These approaches especially probabilistic graphical modeling techniques, which are of great interest to the research community, nicely deal with localness. Typically, these approaches combine first-order logic (to describe the relations (links) between data items) with probabilistic graphical (e.g., Markov logic network [33], Bayesian network [34], inductive logic programming [35]) and machine learning model techniques. In the context of knowledge graph completion, the goal is to predict missing links and assign some probability values to deal with uncertain facts. These probabilistic values assist in determining the correctness of the newly discovered link. One of the examples of these techniques is using Markov Logic Network [33], which, in contrast to existing rule-based approaches, can integrate global information over the knowledge graph. These approaches combine first-order logic with probabilistic methods. The limitation of these techniques is that they are computationally complex and hence are not scalable[36][12].

Probabilistic soft logic [37] is another approach for collective probabilistic reasoning in relational data. It assigns soft-truth values to the assertions/facts from interval [0,1]. In contrast to MLN, it is not computationally complex and provides tractability and scalability. [36]. In [12], the authors propose a knowledge graph identification technique based on Probabilistic soft logic. Extractors assign soft truth values to predicates, and PSL rules are created that run over these statements to perform error correction, deduplication, link prediction, and detecting and removing logical inconsistencies. The refinement scope is determined by the number of PSL rules created before running the process.

*C. Embeddings and Neural network based approaches*

Embeddings-based approaches like TransE [38], TransH [39], TransM [40], TransR [41] have been proved useful, which transforms entities and relationships into their mathematical representation and run extremely fast on GPU/CPU hardware and hence result in a considerably good performance. Besides that, embeddings are used to feed real-valued vectors to train deep neural networks, find similar entities, and enable unsupervised machine learning like clustering. Other techniques to compute embeddings fall in tensor factorization or training the neural network model. Neural tensor models have also recently caught attention. These techniques encode entities and their relationships into low-dimensional vector representations where inference methods can identify new relations. Different embeddings-based techniques for knowledge graph completion [42] have been proposed. For example, [42] has proposed a representation learning framework for property graphs that include rich information in the properties of both nodes and edges. [43] presents an embedding-based link prediction technique that uses PRA for discovering relational paths between entities and then encodes these paths using BiLSTM and CNN. It Combines BiLSTM and CNN with an attention mechanism and captures Semantic correlation between a candidate relation. It can extract Reasoning evidence over multiple paths (not just direct paths) to predict links. The limitation of this approach is that it uses a single type for entity representations and is suited for small KG.

As mentioned previously, embeddings-based techniques cannot represent entities with low degrees. Open-world resources like online-encyclopedias and newswire stories corpus add descriptions to entities to cope with the low-degree issue[44]. Techniques like neighborhood [45][46] and text descriptions [41] have been proposed to handle this issue. For instance, DKRL [41] learns entities' representation not just from TransE [38] but additionally from the textual description of entities mentioned in online encyclopedias. The limitations of these approaches are that they largely depend on the supplement information's availability and quality [22]. For example, ConMask [18] generates embeddings of the entity's names and textual descriptions to find new relations. However, as highlighted in [53], these techniques cannot exploit the rich feature information in the accompanying textual descriptions to detect missing links. [18] uses dependent content masking technique, fully convolutional neural networks, and semantic averaging to extract relationship-dependent embeddings from textual features of entities and relationships. Relationships are discovered based on features mentioned in the content. Nevertheless, it does not provide any mechanism to detect new relations implicitly. Besides, this technique is dependent on the availability and quality of the relationship-dependent text descriptions.

In [44], the authors have highlighted these issues and proposed an MIA (multiple interaction attention) mechanisms to employ a word-level attention mechanism to simulate the interaction between descriptions of entities, relationships, and names of entities and relationships. Also, they use scoring functions to enhance the convergence of the model, which resulted in achieving significant improvements compared to other state-of-the-art models. However, the problem with this approach is that it relies heavily on entity descriptions and only works when necessary information is available.

Wiharja et al., 2020 [15] propose a technique that focuses on generating and correcting statements. It combines different rule-based and embeddings-based (TransE, RUMIS, HAKE, Hermit, etc.) techniques to KG completion iteratively. A reasoner is then used to check the consistency of statements using SHACL and TBox content after each iteration. This scheme's efficacy is limited to the number of SHACL constraints furnished before running consistency checks. Furthermore, it only emphasizes schema-incorrect triples but ignores triples that are not valid but are schema-correct. Another approach that has been proved quite successful in discovering multi-hop relational paths includes reinforcement learning techniques [47][48][49]. These techniques are closely related to embeddings (to achieve better performance and accuracy) and include a reward function that considers different quality parameters to discover links in the knowledge graph. For example, in [50], the authors have proposed a reinforcement learning framework to learn multi-hop relational paths considering accuracy, diversity and efficiency for the reward function. In [51], authors have presented a neural network-based approach which uses a graph convolutional neural network and employs both GAN generative adversial net-based reinforcement learning and LSTM to record trajectory sequence to predict new links.

IV. KG CORRECTNESS APPROACHES

A. Rules and Constraints

Rules and Constraints assist in detecting and resolving contradictory statements and perform the knowledge graph cleaning process [28]. E.g., the Disjoint constraint detects objects that have contradictory types, and sameAs axiom sees duplicate entries. A candidate triple can be scrutinized individually against all the applicable restrictions to check its consistency. Hence the whole knowledge graph is analyzed by scanning each triple against these constraints (ontology axioms) to detect inconsistencies. In [25], the authors have proposed a semi-automatic spotting and resolving of knowledge graph inconsistencies. They utilize a rule-driven methodology that focuses on two causes of inconsistencies, i.e., rules (knowledge graph generation rules) and ontology definitions. They propose a method where they rank the rules and ontology definitions using clustering. Human experts then inspect the top-ranked rules and ontology definitions, and refinements are made. In [52], authors have proposed a Datalog model based on functional dependencies that enable deduction and captures violation of conflicting information. Another scheme is presented in [15], where Ontology axioms and SHACL are used (in the form of constraints) to check the consistency of statements after each iteration of adding new information to the knowledge graph. These constraints must be provided, and there is no mention of how to generate them autonomously. Moreover, it only emphasizes schema-incorrect triples but ignores those triples that are not valid but are schema-correct.

B. Data fusion and Knowledge fusion

Data fusion [53] is another crucial technique to resolve conflicts between different attribute values of the same data item. The objective is to infer the correct value of the data item by engaging multiple sources of information, i.e., correlating, combining, and analyzing values from different data sources to reach an accurate value of the data item. The underlying fusion mechanism usually depends on factors like the reputation/quality of the data source or voting mechanism to support a candidate value based on majority claims [26].

However, in most cases, both approaches are combined to complement each other weaknesses and achieve high accuracy. For example, in [2], authors have presented a technique of fusing different fact extraction methods by combining the signals from each extractor to build a feature vector for each extracted triple. That is how they measure the reliability of each system. They also presented how extractors and priors can be combined to produce high confidence facts. The advanced form of data fusion is knowledge fusion [54]–[56], where the fusion process is targeted towards identifying actual subject-predicate-object values from multiple sources [54]. A set of uncertain statements (subject-predicate-object) extracted from various sources is input to the knowledge fusion process. The validity of the statement and trustworthiness of the data sources are considered and jointly learned into a probability score. In [57], the authors have provided a detailed survey of knowledge fusion techniques to discover and clean errors.

C. Constraint Reasoning and MaxSat

Knowledge fusion and simple constraint checking perform validity on a single statement and can not consider other related statements jointly [26]. For example, if we have multiple statements denoting multiple values for the birthplace of the same person. Each of these statements can be checked and validated individually during constraint checking and could be accepted. The person could have a single birthplace, but which one is true can only be checked jointly through holistic consistency checking. For such situations, different methods exist that take an approach of holistic consistency checking. One of the techniques to implement holistic consistency checking is weighted MaxSat inference. It is an extension of the Maximum Satisfiability problem. This joint inference is performed over the set of weighted candidate triples by imposing soft constraints that can be created manually or learned automatically using pattern mining techniques. These soft constraints carry weights that reflect the confidence score. The goal of the maximum satisfiability problem is to derive as many assertions as trustworthy so that the knowledge graph remains consistent. It ensures that the subset of consistent statements has the accumulative highest weight. Only the minimum set of inconsistent statements will be removed. The main limitation of MaxSat computation is, it is NP-hard and becomes intractable as the number of statements increases.

D. Statistical and Probabilistic approaches

The weighted MaxSat inference mechanism can also be modeled in a probabilistic manner. All the candidate statements and constraints are given as input to the model, and the high-level model is translated into Markov random fields that couples random variables. The Boolean variables become random variables holding probability values for being true or false. To deal with intractability, these models usually make assumptions about conditional independence between random variables. The model is resolved into probabilistic factor graphs, with each factor representing the coupling of subsets of non-independent random variables.

One essential probabilistic graphical modeling technique is the Markov logic network, which enables reasoning over uncertain statements. It applies first-order logic over the Markov network to detect inaccuracies. MLN computes the probability distribution of all worlds, like solving the weighted MaxSat problem, i.e., calculating the world with maximum joint probability. In the MLN world, it is known as MAP inference or maximum a posteriori inference. Just like MaxSat computation, MAP inference is also NP-hard and is not considered for real-world cases. Instead, another technique known as Monte Carlo sampling [33] is adopted for practical cases. For instance, [58] proposes MCMC (i.e., Markov Chain Monte Carlo method) algorithm with a chase-based procedure to deal with approximate marginal inference. They have developed a framework known as soft VADALOG that enables probabilistic reasoning in VADALOG based knowledge graphs. It enables ontological reasoning by supporting features like existential quantification, full recursion and support inductive definitions. It extends VADALOG [59] and Warded Datalog, which supports probabilistic programming and the statistical relational model.

Moreover, MLN can also compute the marginal probability for each random variable which is the same as assigning weight to each candidate statement. However, computing marginal probability using MLN inference is far more computationally complex than MAP inference. So usually, it is not practically used. Furthermore, learning weights for MLN clauses automatically from training data is also a complicated task because it involves non-convex optimization. One of the significant challenges in performing the grounding step (i.e., instantiation of clauses) is that it may result in a combinatorial explosion. In response to this issue, different approaches to lazy grounding techniques have been proposed. For instance, DeepDive [60] project has worked on lazy grounding techniques to avoid unnecessary instantiations [61]. Another approach that is designed to address intractability issues in MLN is PSL (probabilistic soft logic) [36], [62]. Instead of giving Boolean values to variables (i.e., 1 and 0), values ranging between 0 and 1 could be assigned, expressing some degree of belief in a variable. Like MLN, the critical task is to compute the MAP that can become a non-convex operation using the Hinge-loss objective function, making MAP inference non-NP-hard and getting PSL tractable for various real-world use cases.

In [63], the authors propose using PR-OWL-based ontology and MEBN (multi-entity Bayesian network) to model and reason over uncertainty. PR-OWL is a probabilistic extension that supports modeling uncertainty. MEBN, a probabilistic graphical modeling technique, is used to provide reasoning over uncertain knowledge. The approach relies on the manual creation and deletion of ontological as well as for instance-level knowledge. The issue with this scheme is, it does not consider the automatic update of knowledge, and hence updates are made manually.

Similarly, in [62], authors have proposed a probabilistic graphical reasoning framework that detects errors and inconsistencies in an uncertain and temporal knowledge graph. Their focus was to cater to the scalability issues arising from using the Markov logic network. They, therefore, proposed a numerical extension of the Markov logic network that provides a formalism for uncertain and temporal knowledge graphs. Besides, they used a set of Datalog constraints with inequalities to detect inconsistencies.

In [24], authors have presented an effort towards modeling uncertain temporal knowledge graphs. They have proposed a bitemporal model for maintaining and querying uncertain knowledge graphs. This scheme focuses on coalescing (deduplication) under uncertainty. They model temporal facts by combining the transaction time (extraction time) and validity time into the fact. Their approach uses the Markov logic network to model uncertain knowledge graphs and deriving a most probable and consistent temporal knowledge graph. In addition to this, they have devised an algorithm that can scale to compute marginal distributions of temporal queries.

## V. RESEARCH ISSUES AND CHALLENGES

Maintaining coverage, correctness, and freshness of knowledge graphs remains the biggest challenge for industry-scale knowledge graphs. Coverage is still a significant concern, even for huge KBs. Existing knowledge graph construction approaches typically cover basic facts but often fail to cover sophisticated points. Existing knowledge construction schemes primarily focus on knowledge acquisition and lack the capabilities to deal with probabilistic and temporal information. Existing techniques offer clever algorithms and machine learning solutions, each covering a specific aspect of KG construction, but automatically orchestrating and steering the whole construction process remains the challenge.

Based on the Introduction section and evaluation of the literature review, the following research issues have been identified.

- Rule mining and pattern mining approaches provide a convenient and straightforward way of identifying logical patterns and invariants (rules and constraints) that improve KG coverage and correctness. Nevertheless, these approaches suffer from issues like low expressiveness, localness, and strict directionality. The challenge is to devise a tractable rule mining technique to generate expressive soft rules and constraints in OWL-DL using inductive logic programming in dynamic environments.

- Probabilistic approaches to KG refinement simulate human reasoning ability. These methods can model uncertainty and provides holistic consistency checking of KG. But these approaches usually suffer from intractability (for example, MAP reasoning and computing marginal probability are computationally complex problems). Lazy grounding techniques can be devised or extended to address these challenges.

- Distributed representation approaches are scalable and efficiently model the topological characteristics of KG. Nevertheless, they are hard to interpret, cannot model entities with low degrees, suffer from the out-of-vocabulary problem. Moreover, they are not able to effectively encode longer relation paths (semantic relations) between entities. The challenge is to devise a joint framework to create KG embeddings that include intensional and extensional knowledge with a broader range of schema axioms. Moreover, the embeddings should accommodate temporal and probabilistic information to reason over temporal and uncertain facts.

- Neural network-based approaches provide efficient reasoning ability, but their training relies on high computation costs and a large amount of high-quality data. Due to their black-box nature, they are hard to interpret and understand.

## VI. CONCLUSION

Existing techniques for autonomous KG construction have been proved helpful in knowledge collection and harvesting but fail to produce comprehensive knowledge graphs that can support advanced applications like deep question answering and semantic search. This paper will guide finding an efficient solution to this problem by exploring different KG refinement techniques devised to enhance knowledge graphs' overall coverage and correctness and produce knowledge graphs of near-human quality. It discusses various approaches, research issues, merits, and demerits of KG construction techniques, aiming to produce high-quality knowledge graphs, and elaborates on the current development of KG refinement techniques, their strengths, and limitations. We concluded by establishing particular research challenges that need to be addressed to solve the problem.

## REFERENCES

[1] P. Hitzler, "Semantic Web: A Review Of The Field," 2020, doi: 10.1145/3397512.

[2] X. Dong et al., "Knowledge vault: A web-scale approach to probabilistic knowledge fusion," Proc. ACM SIGKDD Int. Conf. Knowl. Discov. Data Min., pp. 601–610, 2014, doi: 10.1145/2623330.2623623.

[3] P. Haase, D. M. Herzig, A. Kozlov, A. Nikolov, and J. Trame, "Metaphactory: A platform for knowledge graph management," Semant. Web, vol. 10, no. 6, pp. 1109–1125, 2019, doi: 10.3233/SW-190360.

[4] B. L. Douglas, "CYC: A Large-Scale Investment in Knowledge Infrastructure," Commun. ACM, vol. 38, no. 11, pp. 33–38, 1995, doi: 10.1145/219717.219745.

[5] R. Speer, J. Chin, and C. Havasi, "ConceptNet 5.5: An Open Multilingual Graph of General Knowledge," AAAI'17 Proc. Thirty-First AAAI Conf. Artif. Intell., no. Singh 2002, pp. 4444–4451, Dec. 2017, [Online]. Available: http://arxiv.org/abs/1612.03975.

[6] D. Vrandečić and M. Krötzsch, "Wikidata: A free collaborative knowledge base," Commun. ACM, vol. 57, no. 10, pp. 78–85, 2014, doi: 10.1145/2629489.

[7] K. Bollacker, C. Evans, P. Paritosh, T. Sturge, and J. Taylor, "Freebase: A collaboratively created graph database for structuring human knowledge," Proc. ACM SIGMOD Int. Conf. Manag. Data, pp. 1247–1249, 2008, doi: 10.1145/1376616.1376746.

[8] J. Weng, Y. Gao, J. Qiu, G. Ding, and H. Zheng, "Construction and Application of Teaching System Based on Crowdsourcing Knowledge Graph," Commun. Comput. Inf. Sci., vol. 1134 CCIS, pp. 25–37, 2019, doi: 10.1007/978-981-15-1956-7_3.

[9] N. Kertkeidkachorn and R. Ichise, "An automatic knowledge graph creation framework from natural language text," IEICE Trans. Inf. Syst., vol. E101D, no. 1, pp. 90–98, 2018, doi: 10.1587/transinf.2017SWP0006.

[10] G. Wu et al., "An Automatic and Rapid Knowledge Graph Construction Method of SG-CIM Model," Proc. - 2020 IEEE Int. Conf. Smart Cloud, SmartCloud 2020, pp. 193–198, 2020, doi: 10.1109/SmartCloud49737.2020.00044.

[11] X. Wu, J. Wu, X. Fu, J. Li, P. Zhou, and X. Jiang, "Automatic knowledge graph construction: A report on the 2019 ICDM/ICBK Contest," Proc. - IEEE Int. Conf. Data Mining, ICDM, vol. 2019-Novem, no. Icdm, pp. 1540–1545, 2019, doi: 10.1109/ICDM.2019.00204.

[12] J. Pujara, H. Miao, L. Getoor, and W. W. Cohen, "Using semantics and statistics to turn data into knowledge," AI Mag., vol. 36, no. 1, pp. 65–74, 2015, doi: 10.1609/aimag.v36i1.2568.

[13] P. Cimiano, "Knowledge Graph Refinement: A Survey of Approaches and Evaluation Methods," Semant. Web, vol. 8, no. 3, pp. 489–508, 2017, [Online]. Available: http://www.semantic-web-journal.net/content/knowledge-graph-refinement-survey-approaches-and-evaluation-methods.

[14] S. Razniewski and P. Das, "Structured Knowledge: Have we made progress? An extrinsic study of KB coverage over 19 years," Int. Conf. Inf. Knowl. Manag. Proc., no. 2, pp. 3317–3320, 2020, doi: 10.1145/3340531.3417447.

[15] K. Wiharja, J. Z. Pan, M. J. Kollingbaum, and Y. Deng, "Schema aware iterative Knowledge Graph completion," J. Web Semant., vol. 65, p. 100616, 2020, doi: 10.1016/j.websem.2020.100616.

[16] X. Chen, S. Jia, and Y. Xiang, "A review: Knowledge reasoning over knowledge graph," Expert Syst. Appl., vol. 141, 2020, doi: 10.1016/j.eswa.2019.112948.

[17] N. Noy, Y. Gao, A. Jain, A. Patterson, A. Narayanan, and J. Taylor, "Industry-scale knowledge graphs lessons and challenges," Queue, vol. 17, no. 2, pp. 1–28, 2019, doi: 10.1145/3329781.3332266.

[18] B. Shi and T. Weninger, "Open-World Knowledge Graph Completion," arXiv, pp. 1957–1964, 2017.

[19] R. West, E. Gabrilovich, K. Murphy, S. Sun, R. Gupta, and D. Lin, "Knowledge base completion via search-based question answering," WWW 2014 - Proc. 23rd Int. Conf. World Wide Web, pp. 515–525, 2014, doi: 10.1145/2566486.2568032.

[20] C. Gutierrez and J. F. Sequeda, "Knowledge Graphs," pp. 3509–3510, 2020, doi: 10.1145/3340531.3412176.

[21] M. Stewart, M. Enkhsaikhan, and W. Liu, "ICDM 2019 knowledge graph contest: Team UWA," Proc. - IEEE Int. Conf. Data Mining, ICDM, vol. 2019-Novem, pp. 1546–1551, 2019, doi: 10.1109/ICDM.2019.00205.

[22] R. Zhang, Y. Mao, and W. Zhao, "Knowledge graphs completion via probabilistic reasoning," Inf. Sci. (Ny)., vol. 521, pp. 144–159, 2020, doi: 10.1016/j.ins.2020.02.016.

[23] E. Huaman, E. Kärle, and D. Fensel, "Knowledge Graph Validation," arXiv, 2020.

[24] M. W. Chekol and H. Stuckenschmidt, "Time-aware probabilistic knowledge graphs," Leibniz Int. Proc. Informatics, LIPIcs, vol. 147, no. 8, pp. 81–817, 2019, doi: 10.4230/LIPIcs.TIME.2019.8.

[25] P. Heyvaert, B. De Meester, A. Dimou, and R. Verborgh, "Rule-driven inconsistency resolution for knowledge graph generation rules," Semant. Web, vol. 10, no. 6, pp. 1071–1086, 2019, doi: 10.3233/SW-190358.

[26] G. Weikum, S. Razniewski, L. Dong, and F. Suchanek, "Machine knowledge: Creation and curation of comprehensive knowledge bases," arXiv, 2020.

[27] N. Kertkeidkachorn and R. Ichise, "T2KG: An end-to-end system for creating knowledge graph from unstructured text," AAAI Work. - Tech. Rep., vol. WS-17-01-, pp. 743–749, 2017.

[28] N. Ahmadi, V. P. Huynh, V. Meduri, S. Ortona, and P. Papotti, "Mining Expressive Rules in Knowledge Graphs," J. Data Inf. Qual., vol. 12, no. 2, 2020, doi: 10.1145/3371315.

[29] A. C. Kakas, R. A. Kowalski, and F. Toni, "Inductive logic programming," J. Log. Comput., vol. 2, no. 6, pp. 719–770, 1992, doi: 10.1093/logcom/2.6.719.

[30] L. Galárraga, C. Teflioudi, K. Hose, and F. M. Suchanek, "AMIE: Association rule mining under incomplete evidence in ontological knowledge bases," WWW 2013 - Proc. 22nd Int. Conf. World Wide Web, pp. 413–422, 2013.

[31] Y. Chen, D. Z. Wang, and S. Goldberg, "ScaLeKB: scalable learning and inference over large knowledge bases," VLDB J., vol. 25, no. 6, pp. 893–918, 2016, doi: 10.1007/s00778-016-0444-3.

[32] M. Nickel, K. Murphy, V. Tresp, and E. Gabrilovich, "A review of relational machine learning for knowledge graphs," Proc. IEEE, vol. 104, no. 1, pp. 11–33, 2016, doi: 10.1109/JPROC.2015.2483592.

[33] M. Richardson and P. Domingos, "Markov logic networks," Mach. Learn., vol. 62, no. 1-2 SPEC. ISS., pp. 107–136, 2006, doi: 10.1007/s10994-006-5833-1.

[34] G. F. Cooper and E. Herskovits, "A Bayesian Method for the Induction of Probabilistic Networks from Data," Mach. Learn., vol. 9, no. 4, pp. 309–347, 1992, doi: 10.1023/A:1022649401552.

[35] S. Muggleton and L. De Raedt, "Inductive Logic Programming: Theory and Methods," Arch. Pediatr. Adolesc. Med., vol. 148, no. 2, pp. 209–210, 1994, doi: 10.1001/archpedi.1994.02170020095018.

[36] J. Pujara, H. Miao, L. Getoor, and W. Cohen, "Knowledge graph identification," Lect. Notes Comput. Sci. (including Subser. Lect. Notes Artif. Intell. Lect. Notes Bioinformatics), vol. 8218 LNCS, no. PART 1, pp. 542–557, 2013, doi: 10.1007/978-3-642-41335-3_34.

[37] A. Kimmig, S. H. Bach, M. Broecheler, B. Huang, and L. Getoor, "A Short Introduction to Probabilistic Soft Logic," Proc. NIPS Work. Probabilistic Program. Found. Appl., no. 1, pp. 1–4, 2012.

[38] A. Bordes et al., "Translating Embeddings for Modeling Multi-relational Data To cite this version : HAL Id : hal-00920777 Translating Embeddings for Modeling Multi-relational Data," 2013.

[39] Z. Wang, J. Zhang, J. Feng, and Z. Chen, "Knowledge graph embedding by translating on hyperplanes," Proc. Natl. Conf. Artif. Intell., vol. 2, pp. 1112–1119, 2014.

[40] M. Fan, Q. Zhou, E. Chang, and T. F. Zheng, "Transition-based knowledge graph embedding with relational mapping properties," Proc. 28th Pacific Asia Conf. Lang. Inf. Comput. PACLIC 2014, pp. 328–337, 2014.

[41] Y. Lin, Z. Liu, M. Sun, Y. Liu, and X. Zhu, "Learning entity and relation embeddings for knowledge graph completion," Proc. Natl. Conf. Artif. Intell., vol. 3, pp. 2181–2187, 2015.

[42] Y. Hou, H. Chen, C. Li, J. Cheng, and M. C. Yang, "A representation learning framework for property graphs," Proc. ACM SIGKDD Int. Conf. Knowl. Discov. Data Min., pp. 65–73, 2019, doi: 10.1145/3292500.3330948.

[43] B. Jagvaral, W. K. Lee, J. S. Roh, M. S. Kim, and Y. T. Park, "Path-based reasoning approach for knowledge graph completion using CNN-BiLSTM with attention mechanism," Expert Syst. Appl., vol. 142, p. 112960, 2020, doi: 10.1016/j.eswa.2019.112960.

[44] L. Niu et al., "Open-world knowledge graph completion with multiple interaction attention," World Wide Web, vol. 24, no. 1, pp. 419–439, 2021, doi: 10.1007/s11280-020-00847-2.

[45] M. Schlichtkrull, T. N. Kipf, P. Bloem, R. van den Berg, I. Titov, and M. Welling, "Modeling Relational Data with Graph Convolutional Networks," Lect. Notes Comput. Sci. (including Subser. Lect. Notes Artif. Intell. Lect. Notes Bioinformatics), vol. 10843 LNCS, no. 1, pp. 593–607, 2018, doi: 10.1007/978-3-319-93417-4_38.

[46] F. Kong, R. Zhang, Y. Mao, and T. Deng, "LENA: Locality-expanded neural embedding for knowledge base completion," 33rd AAAI Conf. Artif. Intell. AAAI 2019, 31st Innov. Appl. Artif. Intell. Conf. IAAI 2019 9th AAAI Symp. Educ. Adv. Artif. Intell. EAAI 2019, pp. 2895–2902, 2019, doi: 10.1609/aaai.v33i01.33012895.

[47] Xiao Lin, P. Subasic, and H. Yin, "Rel4KC: A Reinforcement Learning Agent for Knowledge Graph Completion and Validation," Annu. Conf. Innov. Technol. Comput. Sci. Educ. ITiCSE, p. 1291, 2020.

[48] P. Tiwari, H. Zhu, and H. M. Pandey, "DAPath: Distance-aware knowledge graph reasoning based on deep reinforcement learning," Neural Networks, vol. 135, pp. 1–12, 2021, doi: 10.1016/j.neunet.2020.11.012.

[49] G. Wan, S. Pan, C. Gong, C. Zhou, and G. Haffari, "Reasoning like human: Hierarchical reinforcement learning for knowledge graph reasoning," IJCAI Int. Jt. Conf. Artif. Intell., vol. 2021-Janua, pp. 1926–1932, 2020, doi: 10.24963/ijcai.2020/267.

[50] W. Xiong, T. Hoang, and W. Y. Wang, "DeepPath: A reinforcement learning method for knowledge graph reasoning," EMNLP 2017 - Conf. Empir. Methods Nat. Lang. Process. Proc., pp. 564–573, 2017, doi: 10.18653/v1/d17-1060.

[51] Q. Wang, Y. Ji, Y. Hao, and J. Cao, "GRL: Knowledge graph completion with GAN-based reinforcement learning," Knowledge-Based Syst., vol. 209, p. 106421, 2020, doi: 10.1016/j.knosys.2020.106421.

[52] S. Abiteboul et al., "Deduction with Contradictions in Datalog To cite this version : Deduction with Contradictions in Datalog," 2014.

[53] J. Bleiholder and F. Naumann, "Data Fusion," ACM Comput. Surv., vol. 41, no. 1, pp. 1–41, 2009, doi: 10.1145/1456650.1456651.

[54] X. L. Dong et al., "From data fusion to knowledge fusion," Proc. VLDB Endow., vol. 7, no. 10, pp. 881–892, 2014, doi: 10.14778/2732951.2732962.

[55] A. Smirnov and T. Levashova, "Knowledge fusion patterns: A survey," Inf. Fusion, vol. 52, no. June 2018, pp. 31–40, 2019, doi: 10.1016/j.inffus.2018.11.007.

[56] X. Zhao, Y. Jia, A. Li, R. Jiang, and Y. Song, "Multi-source knowledge fusion: a survey," World Wide Web, vol. 23, no. 4, pp. 2567–2592, 2020, doi: 10.1007/s11280-020-00811-0.

[57] X. L. Dong and D. Srivastava, "Knowledge curation and knowledge fusion: Challenges, models, and applications," Proc. ACM SIGMOD Int. Conf. Manag. Data, vol. 2015-May, pp. 2063–2066, 2015, doi: 10.1145/2723372.2731083.

[58] L. B. B, E. Laurenza, and E. Sallinger, Reasoning under uncertainity in Knowledge Graphs, vol. 1. Springer International Publishing, 2020.

[59] L. Bellomarini, D. Benedetto, G. Gottlob, and E. Sallinger, "Vadalog: A modern architecture for automated reasoning with large knowledge graphs," Inf. Syst., no. xxxx, p. 101528, 2020, doi: 10.1016/j.is.2020.101528.

[60] F. Niu, C. Zhang, C. Ré, and J. Shavlik, "DeepDive: Web-scale knowledge-base construction using statistical learning and inference," CEUR Workshop Proc., vol. 884, pp. 25–28, 2012.

[61] F. Niu, C. Ré, A. H. Doan, and J. Shavlik, "Tuffy: Scaling up statistical inference in markov logic networks using an RDBMS," Proc. VLDB Endow., vol. 4, no. 6, pp. 373–384, 2011, doi: 10.14778/1978665.1978669.

[62] M. W. Chekol, G. Pirrò, J. Schoenfisch, and H. Stuckenschmidt, "Marrying uncertainty and time in knowledge graphs," 31st AAAI Conf. Artif. Intell. AAAI 2017, pp. 88–94, 2017.

[63] S. Wang, Y. Zhang, and Z. Liao, "Building Domain-Specific Knowledge Graph for Unmanned Combat Vehicle Decision Making under Uncertainty," Proc. - 2019 Chinese Autom. Congr. CAC 2019, pp. 4718–4721, 2019, doi: 10.1109/CAC48633.2019.8996418.